\documentclass[10pt,conference,a4paper]{IEEEtran}

\usepackage{graphicx}
\usepackage{framed,multirow}
\usepackage{amssymb}
\usepackage{latexsym}
\usepackage{amsmath}
\usepackage{balance}
\usepackage{geometry}

\newtheorem{theorem}{Theorem}

\newtheorem{lemma}{Lemma}

\usepackage{url}
\usepackage{xcolor}
\definecolor{newcolor}{rgb}{.8,.349,.1}

\usepackage{color}

\begin{document}

\title{Revealing Structure in Large Graphs:\\
Szemer\'edi's Regularity Lemma and Its Use in Pattern Recognition}

\author{\IEEEauthorblockN{\Large{Marcello Pelillo}}
\IEEEauthorblockA{ECLT, Ca' Foscari University \\ S. Marco 2940, Venice, Italy \\ DAIS, Ca' Foscari University\\ Via Torino 155, Venice, Italy\\
Email: pelillo@unive.it}
\and

\IEEEauthorblockN{\Large{Ismail Elezi}}
\IEEEauthorblockA{DAIS, Ca' Foscari University\\ Via Torino 155, Venice, Italy\\
Email: ismail.elezi@gmail.com}
\and

\IEEEauthorblockN{\Large{Marco Fiorucci}}
\IEEEauthorblockA{DAIS, Ca' Foscari University\\ Via Torino 155, Venice, Italy\\
Email: marco.fiorucci@unive.it}}

\date{}

\maketitle

\begin{abstract}
Introduced in the mid-1970's as an intermediate step in proving a long-standing conjecture on arithmetic progressions, Szemer\'edi's regularity lemma has emerged over time as a fundamental tool in different branches of graph theory, combinatorics and theoretical computer science. Roughly, it states that every graph can be approximated by the union of a small number of random-like bipartite graphs called regular pairs. In other words, the result provides us a way to obtain a good description of a large graph using a small amount of data, and can be regarded as a manifestation of the all-pervading dichotomy between structure and randomness. 
In this paper we will provide an overview of the regularity lemma and its algorithmic aspects, and will discuss its relevance in the context of pattern recognition research.
\end{abstract}

%\linenumbers

%% main text
\section{Introduction}
\label{sec1}

In 1941, the Hungarian mathematician P. Tur\'an provided an answer to the following innocent-looking question. What is the maximal number of edges in a graph with $n$ vertices not containing a complete subgraph of order $k$, for a given $k$? This graph is now known as a Tur\'an graph and contains no more than $n^2(k-2)/2(k-1)$ edges. Later, in another classical paper, T. S. Motzkin and E. G. Straus (1965)\nocite{MotStr65} provided a novel proof of Tur\'an's theorem using a continuous characterization of the clique number of a graph. Thanks to contributions of P. Erd\"os, B. Bollob\'as, M. Simonovits, E. Szemer\'edi and others, Tur\'an's study developed soon into one of the richest branches of 20th-century graph theory, known as \textit{extremal graph theory}, which has intriguing connections with Ramsey theory, random graph theory, algebraic constructions, etc. Roughly, extremal graph theory studies how the intrinsic structure of graphs ensures certain types of properties (e.g., cliques, coloring and spanning subgraphs) under appropriate conditions (e.g., edge density and minimum degree) \cite{Bol04}.

Among the many achievements of extremal graph theory, Szemer\'edi's regularity lemma is certainly one of the best known \cite{Die10}. Basically, it states that every graph can be partitioned into a small number of random-like bipartite graphs, called regular pairs, and a few leftover edges. Szemer\'edi's result was introduced in the mid-seventies as an auxiliary tool for proving the celebrated Erd\"os-Tur\'an conjecture on arithmetic progressions in dense sets of integers \cite{Sze75}.
Over the past two decades, this result has been refined, extended and interpreted in several ways and has now become an indispensable tool in discrete mathematics and theoretical computer science \cite{KomSim96,Kom+02,Tao06,LovSze07}. 
Interestingly, an intriguing connection has also been established between the (effective) testability of graph properties (namely, properties that are testable with a constant number of queries on a graph) and regular partitions \cite{Alo+09}.
These results provide essentially a way to obtain a good description of a large graph using a small amount of data, and can be regarded as a manifestation of the all-pervading dichotomy between structure and randomness.

Indeed, the notion of separating structure from randomness in large (and possibly dynamic) data sets is prevalent in nearly all domains of applied science and technology, as evidenced by the importance and ubiquity of clustering methods in data analysis. However, unlike standard clustering approaches, regular partitions minimize discrepancies both within and between clusters in the sense that the members of a cluster behave roughly similarly toward members of each (other or own) cluster \cite{Bol16,Alo+10}. This is a new paradigm for structural decomposition, which distinguishes it radically from all prior works in data analysis. This property allows for exchangeability among members of distinct parts within the partition,
which can be important in a variety of real-world scenarios.\footnote{We are grateful to Marianna Bolla for stressing this point in a personal communication.}

In its original form, the regularity lemma is an existence predicate. Szemer\'edi demonstrated the existence of a regular partition under the most general conditions, but he did not provide a constructive way to obtain it. However, in the mid-1990's, \cite{Alo+94} succeeded in developing the first algorithm to create a regular partition on arbitrary graphs, and showed that it has polynomial computational complexity. Other polynomial-time algorithms can be found, e.g., in \cite{FriKan99,CzyRod00,KohRodTho03}.

In this paper we will provide an overview of the regularity lemma and its algorithmic aspects, and will discuss its relevance in the context of structural pattern recognition. We will focus, in particular, on graph-based clustering and image segmentation, and will show how the notion of a regular partition and
associated algorithms can provide fresh insights into old pattern recognition and machine learning problems.
Preliminary results on some real-world data seem to suggest that, although Szemer\'edi's lemma is a result concerning very large graphs, 'regular-like' structures can appear already in suprisingly small-scale graphs. The strength of regular partitions, though, is expected to reveal itself in larger and larger graphs and, if confirmed, this would pave the way for a principled approach to big data analysis. Presently, virtually all standard methods for dealing with big data are based on classical clustering techniques, such as $k$-means or variations thereof. Regular-like partitions could offer a different, more principled perspective to the problem by providing more informative structures than traditional clustering methods.

\section{Szemer\'edi's regularity lemma}
\label{sec:lemma}

Let $G = (V,E)$ be an undirected graph with no self-loops, where $V$ is the set of vertices and $E$ is the set of edges, and let $ X, Y \subseteq V $ be two disjoint subsets of vertices of $G$. We define the {\em edge density} of the pair $(X,Y)$ as:
\begin{equation}
\label{density}
d(X,Y) = \frac{e(X,Y)}{\left\vert{X}\right\vert \left\vert{Y}\right\vert}
\end{equation}
where $e(X,Y)$ denotes the number of edges of $G$ with an endpoint in $X$ and an endpoint in $Y$, and $\left\vert{~\cdot~}\right\vert $ denotes the cardinality of a set. The edge densities are real numbers between 0 and 1.

Given a positive constant $\varepsilon > 0$, we say that the pair $(A,B)$ of disjoint vertex sets $ A,B \subseteq V $ is {\em $\varepsilon$-regular}
if for every $ X \subseteq A $ and $ Y \subseteq B $ satisfying
\begin{equation}
\left\vert{X}\right\vert > \varepsilon \left\vert{A}\right\vert \hspace{.5cm} \mbox{and} \hspace{.5cm} \left\vert{Y}\right\vert > \varepsilon \left\vert{B}\right\vert
\end{equation}
we have
\begin{equation}
\left\vert{d(X,Y) - d(A,B)}\right\vert < \varepsilon~.
\end{equation}
Thus, in an $\varepsilon$-regular pair the edges are distributed fairly uniformly.

A partition of $V$ into pairwise disjoint classes $C_0, C_1, . . . , C_k$ is called {\em equitable} if all the classes $C_i$ ($1 \leq i \leq k$) have the same cardinality. The {\em exceptional} set $C_0$ (which may be empty) has only a technical purpose: it makes it possible that all other classes have exactly the same number of vertices. 

An equitable partition $C_0, C_1, . . . , C_k$, with $C_0$ being the exceptional set, is called {\em $\varepsilon$-regular} if:
\begin{enumerate}
\item $|C_0| < \varepsilon|V |$ %%% and
\item all but at most $\varepsilon$$k^2$ of the pairs $(C_i, C_j)$  are $\varepsilon$-regular ($1 \leq i < j \leq k$)
\end{enumerate}

\begin{theorem}[Szemer\'edi's regularity lemma (1976)]
For every positive real $\varepsilon$ and for every positive integer $m$, there are positive integers $N = N(\varepsilon,m)$ and $M = M(\varepsilon,m)$ with the following property: for every graph $G=(V,E)$, with $\left\vert{V}\right\vert \geq N$, there is an $\varepsilon$-regular partition of $G$ into $k + 1$ classes such that $m \leq k \leq M$.
\end{theorem}

Given an $r \times r$ symmetric matrix $(p_{ij})$ with $0 \leq p_{ij} \leq 1$, and positive integers $n_1, n_2, . . . , n_r$, a generalized random graph $R_n$ for $n = n_1 + n_2 + . . . + n_r$ is obtained by partitioning $n$ vertices into classes $C_i$ of size $n_i$ and joining the
vertices $x \in V_i$, $y \in V_j$ with probability $p_{ij}$, independently for all pairs $\{x, y\}$. Now, as pointed out by
Koml\'os and Simonovits (1996)\nocite{KomSim96}, the regularity lemma asserts basically that every graph can be approximated by generalized random graphs. Note that, for the lemma to be useful, the graph has to to be dense. Indeed, for sparse graphs it becomes trivial as all densities of pairs tend to zero \cite{Die10}.

The lemma allows us to specify a lower bound $m$ on the number of classes.
A large value of $m$ ensures that the partition classes $C_i$ are sufficiently small, thereby increasing the proportion of (inter-class) edges subject to the regularity condition and reducing the intra-class ones. The upper bound $M$ on the number of partitions guarantees that for large graphs the partition sets are large too. Finally, it should be noted that a singleton partition is $\varepsilon$-regular for every value
of $\varepsilon$ and $m$.

%The regularity lemma permits $\varepsilon$ $k^2$ pairs to be irregular. Following a path pointed out by Szemer\'edi [16], many researchers studied if the result could be strengthened, avoiding the presence of such pairs. However, it turned out that forcing the lemma in that way would affect its generality [1]. 

The problem of checking if a given partition is $\varepsilon$-regular is a quite surprising
one from a computational complexity point of view. In fact, it turns out that constructing an $\varepsilon$-regular partition is easier than checking if a given one responds to the $\varepsilon$-regularity criterion.

\begin{theorem}[Alon et al., 1994]
The following decision problem is co-NP-complete. Given a graph $G$, an integer $k \geq 1$, a parameter $\varepsilon > 0$, 
and a partition of the set of vertices of $G$ into $k + 1$ parts. Decide if the given partition is $\varepsilon$-regular.
\end{theorem}

Recall that the complexity class {\em co-NP-complete} collects all problems whose complements are NP-complete. 
In other words, a co-NP-complete problem is one for which there are efficiently verifiable proofs of its no-instances, i.e., its counterexamples.

Before concluding this section, we mention that after the publication of Szemer\'edi's original lemma
several variations, extensions and interpretations have been proposed in the literature.
In particular, we have got weaker regularity notions \cite{FriKan99:weak,LovSze07}
and stronger ones \cite{Alo+00,Tao06,LovSze07}, and we have also got versions for sparse graphs and matrices
\cite{GerSte05,Sco11} and hypergraphs \cite{CzyRod00,FraRod02}. Interestingly, \cite{Tao06} provided an interpretation of the lemma in terms of information theory, while \cite{LovSze07} offered 
three different analytic interpretations.

%%%%%%%% VARIATIONS AND EXTENSIONS
%?Weak? regularity lemma (Frieze and Kannan, 1996, 1999)
%?Strong? regularity lemma (Alon, Fisher, Krivelevich and Szegedy, 2000; Tao, 2006; Lov�sz and Szegedy 2007)
%Sparse graphs (Kohayakawa, 1993; R�dl, 1993; Scott, 2010)
%Hypergraphs (Czygrinow and R�dl, 2000; Frankl and R�dl, 2002; R�dl and Skokan, 2004; Tao, 2006; Gowers, 2007) 
%Information-theoretic interpretation (Tao, 2005)
%Analytic interpretation (Lov�sz and Szegedy, 2006)

%%%% APPLICATIONS
%Szemeredi?s regularity lemma has found countless applications in many areas of mathematics, such as:
%Number theory
%Extremal graph theory
%Ramsey theory (Chvatal et al., 1983)
%Property testing (Goldreich, Goldwasser and Ron, 1998)
%Approximation algorithms (Frieze and Kannan, 1999)
%Graph limits (Lov�sz, 2012)

\section{Finding regular partitions}
\label{sec:algo}

The original proof of the regularity lemma \cite{Sze76} is not constructive, yet this has not narrowed the range of its applications in such fields as extremal graph theory, number theory and combinatorics.
However, \cite{Alo+94} proposed a new formulation of the lemma which emphasizes the algorithmic nature of the result.

\begin{theorem}[Alon et al., 1994]
For every $\varepsilon > 0$ and every positive integer t there is an integer $Q = Q(\varepsilon,t) $ such that every graph with $n > Q$ vertices has an $\varepsilon$-regular partition into $k+1$ classes, where $t \leq k \leq Q$. For every fixed $\varepsilon > 0$ and $t \geq 1$ such a partition can be found in $O(M(n))$ sequential time, where $M(n) = O(n^{2.376})$ is the time for multiplying two $n \times n$ matrices with 0,1 entries over the integers. It can also be found in time $O(log n)$ on an EREW PRAM with a polynomial number of parallel processors.
\end{theorem}

In the remaining of this section we shall derive the algorithm alluded to in the previous theorem.
We refer the reader to the original paper \cite{Alo+94} for more technical details and proofs.

Let $H$ be a bipartite graph with equal color classes
$\left\vert{A}\right\vert = \left\vert{B}\right\vert = n$.
We define the {\em average degree} of $H$ as
\begin{equation}
d = \frac{1}{2n} \sum_{i \in A \cup B} deg(i)
\end{equation}
where $deg(i)$ is the the degree of vertex $i$.

For two distinct vertices $y_1, y_2 \in B$ define the {\em neighbourhood deviation} of $y_1$ and $y_2$ by
\begin{equation}
\sigma(y_1,y_2) = \left\vert{N(y_1) \cap N(y_2)}\right\vert - \frac{d^2}{n}
\end{equation}
where $N(x)$ is the set of neighbors of vertex $x$.
For a subset $Y \subseteq B$ the deviation of $Y$ is defined as:
\begin{equation}
\sigma(Y)  = \frac{\sum_{y_1,y_2 \in Y} \sigma(y_1,y_2)} {\left\vert{Y}\right\vert^2}
\end{equation}

Now, let $0 < \varepsilon < 1/16$. It can be shown that if there exists $Y \subseteq B, \left\vert{Y}\right\vert > \varepsilon n$
such that $\sigma(Y ) \geq  {\varepsilon^3 n / 2}$, then at least one of the following cases occurs:

\begin{enumerate}

\item
$d < \varepsilon^3 n$ (which amounts to saying that $H$ is $\varepsilon$-regular); 

\item
there exists in $B$ a set of more than $\frac{1}{8} \varepsilon^4 n$ vertices whose degree deviate from $d$ by at least $\varepsilon^4 n$;

\item
there are subsets $A' \subset A$, $B' \subset B$, $\left\vert{A'}\right\vert \geq \frac{\varepsilon^4}{4}n$, $\left\vert{B'}\right\vert \geq \frac{\varepsilon^4}{4}n$ and $ \left\vert{d(A',B') - d(A,B)}\right\vert \geq \varepsilon^4$.

\end{enumerate}

Note that one can easily check if 1 holds in time $O(n^2)$. Similarly, it is trivial to check if 2 holds in $O(n^2)$ time, and in case it holds to exhibit the required subset of $B$ establishing this fact. 
If both cases above fail we can proceed as follows.
For each $y_0 \in B$ with $\left\vert{deg(y) - d}\right\vert < \varepsilon^4 n$
we find the set
$B_{y_0} = \{~ y \in B ~:~ \sigma(y_0,y) \geq 2 \varepsilon^4 n/ 4 ~\}$.
It can be shown that there exists at least one such $y_0$ for which $\left\vert{B_{y_0}}\right\vert \geq \varepsilon^4 n$. 
The subsets $B' = B_{y_0}$ and $A' = N(y_0)$ are the
required ones. Since the computation of the quantities $\sigma(y, y')$, for $y, y' \in B$, can be done by squaring the adjacency matrix of $H$, the overall complexity of this algorithms is $O(M(n)) = O(n^{2.376})$.

In order to understand the final partitioning algorithm we need the following two lemmas.

\begin{lemma}[Alon et al., 1994]
Let $H$ be a bipartite graph with equal classes $\left\vert{A}\right\vert = \left\vert{B}\right\vert = n$. Let $2n^{-\frac{1}{4}} < \varepsilon < \frac{1}{16}$. 
There is an $O(n^{2.376})$ algorithm which verifies that $H$ is $\varepsilon$-regular or finds two
subsets $A' \subseteq A$ and $B' \subseteq B$ such that 
$|A'| \geq \frac{\varepsilon^4}{4}n$,
$|B'| \geq \frac{\varepsilon^4}{4}n$, and
$|d(A',B')-d(A,B)|\geq \varepsilon ^4$.
%Assume that at most $\frac{\varepsilon^4 n}{8}$ vertices of B deviate from the average degree of $H$ by at least $\varepsilon^4 n$. Then, if $H$ is not $\varepsilon$-regular there exists $Y \cup B$, $\left\vert{Y}\right\vert \geq \varepsilon n$ such that $\sigma(Y) \geq \frac{\varepsilon^3 n}{2}$.
\end{lemma}

It is quite easy to check that the regularity condition can be rephrased in terms of the average degree of $H$. Indeed, it can be seen that if $d < \varepsilon^3n$, then $H$ is $\varepsilon$-regular, and this can be tested in $O(n^2)$ time. Next, it is necessary to count the number of vertices in $B$ whose degrees deviate from $d$ by at least $\varepsilon^4n$. Again, this operation takes $O(n^2)$ time. If the number of deviating vertices is more than $\frac{\varepsilon^4n}{8}$, then the degrees of at least half of them deviate in the same direction and if we let $B'$ be such a set of vertices and $A' = A$ we are done. Otherwise, it can be shown that there must exist $Y \subseteq B$ such that $\left\vert{Y}\right\vert \geq \varepsilon n$ and $\sigma(Y) \geq \frac{\varepsilon^3n}{2}$. Hence, our previous discussion shows that the required subsets $A'$ and $B'$ can be found in $O(n^{2.376})$ time.

Given an equitable partition $P$ of a graph $G = (V,E)$ into classes $C_0, C_1 . . . C_k,$ 
\cite{Sze76} defines a measure called {\em index of partition}:
\begin{equation}
\label{indexP}
ind(P) = \frac{1}{k^2} \sum_{s=1}^{k} \sum_{t=s+1}^{k} d(C_s,C_t)^2~.
\end{equation}
Since $0 \leq d(C_s, C_t) \leq 1$, $1 \leq s$, $t \leq k$, it can be seen that $ind(P) \leq 1/2$.

The following lemma is the core of Szemer\'edi's original proof.

\begin{lemma}[Szemer\'edi, 1976]
Fix k and $\gamma$ and let $G = (V,E)$ be a graph with n vertices. Let P be an equitable partition of V into classes $C_0,C_1,...,C_k.$. Assume $ \left\vert{C_1}\right\vert > 4^{2k}$ and $4^k > 600 \gamma^{-5}$. Given proofs that more than $\gamma k^2$ pairs $(C_r,C_s)$ are not $\gamma$-regular, then one can find in $O(n)$ time a partition P' (which is a refinement of P) into $1+k4^k$ classes, with the exceptional class of cardinality at most $\left\vert{C_0}\right\vert + \frac{n}{4^k}$ and such that
\begin{equation}
ind(P')  \geq ind(P) + \frac{\gamma^5}{20}.
\end{equation}
\end{lemma}

The idea formalized in the previous lemma is that, if a partition violates the regularity condition, then it can be refined by a new partition and, in this case, the index of partition measure can be improved. On the other hand, the new partition adds only few elements to the current exceptional set so that, in the end, its cardinality will respect the definition of equitable partition.

We are now in a position to sketch the complete partitioning algorithm. The procedure is divided into two main steps: in the first step all the constants needed during the next computation are set; in the second one, the partition is iteratively created. An iteration is called \textit{refinement step}, because, at each iteration, the current partition is closer to a regular one.

Given any $\varepsilon > 0$ and a positive integer $t$, we define the constants $N = N(\varepsilon,t)$ and $T = T(\varepsilon,t)$ as follows; let $b$ be the least positive integer such that
\begin{equation}
4^b > 600 (\frac{\varepsilon^4}{16})^{-5},   b \geq t.
\end{equation}
Let $f$ be the integer valued function defined inductively as 
\begin{equation}
f(0) = b, f(i+1) = f(i)4^{f(i)}.
\end{equation}
Put $T = f(\lceil10 (\frac{\varepsilon^4}{16})^{-5}\rceil)$ and $N = max\{T4^{2T}, \frac{32T}{\varepsilon^5}\}$. 
The algorithm for a graph $G=(V,E)$ with $n$ vertices is as follows:
\begin{enumerate}

\item
{\bf Create the initial partition:}
Arbitrarily divide the vertices of $G$ into an equitable partition $P_1$ with classes $C_0, C_1,...,C_b$ where $\left\vert{C_1}\right\vert = \lfloor{\frac{n}{b}}\rfloor$ and hence $\left\vert{C_0}\right\vert < b$. Denote $k_1 = b$ 

\item
{\bf Check regularity:}
For every pair $(C_r,C_s)$ of $P_i$, verify if it is $\varepsilon$-regular or find $X \subseteq C_r$, $Y \subseteq C_s$, $\left\vert{X}\right\vert \geq \frac{\varepsilon^4}{16} \left\vert{C_1}\right\vert$, $\left\vert{Y}\right\vert \geq \frac{\varepsilon^4}{16} \left\vert{C_1}\right\vert$, such that $\left\vert{d(X,Y) - d(C_s, C_t}\right\vert \geq \varepsilon^4$

\item
{\bf Count regular pairs:}
If there are at most $\varepsilon \binom{k_i}{2}$ pairs that are not verified as $\varepsilon$-regular, then halt. $P_i$ is an $\varepsilon$-regular partition

\item
{\bf Refine:}
Apply the refinement algorithm (Lemma 2) where $P=P_i$, $k = k_i$, $\gamma = \frac{\varepsilon^4}{16}$ and obtain a partition $P'$ with $1+k_i 4^{k_i}$ classes

\item
{\bf Iterate:}
Let $k_{i+1} = k_i 4{k_i}$, $P_{i+1} = P'$, $i=i+1$, and go to step (2)
\end{enumerate}

%%%Later this algorithm was improved by Kohayakawa, Rodl, and Thoma [11] who gave a deterministic algorithm for finding a 
%%%regular partition in time $O(n^2)$.

The algorithm described above for finding a regular partition was the first one proposed in the literature.
Later, other algorithms have been developed which improve the original one in several respects.
In particular, we mention an algorithm developed by Frieze and Kannan (1999)\nocite{FriKan99},
which is based on an intriguing relation between the regularity conditions and the singular values of
matrices, and Czygrinow and R\"odl's (2000)\nocite{CzyRod00}, who proposed a new algorithmic version
of Szemer\'edi's lemma for hypergraphs. The latter paper also showed how to deal with edge-weighted graphs.
To do so, we simply need to replace the original notion of a density introduced in (\ref{density}), with
the following one:
\begin{equation}
\label{weighted-density}
 d_w(X,Y) = \frac{\sum_{i=1}^{\left\vert{X}\right\vert} \sum_{j=1}^{\left\vert{Y}\right\vert} w(x_i, y_j)}{\left\vert{X}\right\vert \left\vert{Y}\right\vert}
\end{equation}
which takes into account the edge weights.
More recent algorithms for finding regular partitions can be found in \cite{KohRodTho03,FisMatSha07,Del+12}.

The algorithmic solutions developed so far have been focused exclusively on {\em exact} algorithms whose worst-case complexity, although being polynomial in the size of the underlying graph, has a hidden tower-type dependence on an accuracy parameter. In fact, \cite{Gow97} proved that this tower function is necessary in order to guarantee a regular partition for {\em all} graphs. This has typically discouraged researchers from applying regular partitions to practical problems, thereby confining them to the purely theoretical realm.  

To make the algorithm truly applicable, \cite{SpePel07}, and later \cite{Sar+12},
instead of insisting on provably regular partitions, proposed a few simple heuristics that try to 
construct an approximately regular partition. 
In particular, the main obstacle towards a practical version of the algorithm is the refinement step, which
involves the creation of an exponentially large number of subclasses at each iteration.
Indeed, Step 2 of the original algorithm finds all possible irregular pairs in the graph. As a consequence, each class may be involved with up to $(k-1)$ irregular pairs,
$k$ being the number of classes in the current partition, thereby leading to an exponential growth.
To avoid the problem, for each class, one can limit the number of irregular pairs containing it to at most one, possibly chosen randomly among all irregular pairs.
This simple modification allows one to divide the classes into a constant, rather than exponential, number of subclasses $l$ (typically $ 2 \leq l \leq 7$).
Despite the crude approximation this seems to work well in practice.
We refer the reader to the original papers for more details.

%The final partition found in this way behaves just like a regular partition (especially forgraphs appearing in practice) and yet it does not require the large number of vertices as required by the original lemma. 

%In particular, they solved the exponential growth, by dividing the atoms in each iteration by a number $l$ (typically $ 2 \leq l \leq 7$) instead of dividing by $k4^k$ (k is the number of classes in the previous iteration). While there aren't theoretical guarantees, this partition behaves just like a regular partition (especially for graphs appearing in practice) and yet it does not require the large number of vertices as required by the original lemma. 

\section{Using the regularity lemma for pairwise clustering}
\label{sec:clustering}

\cite{SpePel07} reported arguably the first practical application of the regularity lemma and related algorithms. The original motivation was to study how to take advantage of the information
provided by Szemer\'edi's partitions in a pairwise clustering context. 
Here, the regularity lemma is used as a preclustering strategy, in an attempt to work on a more compact, yet informative, structure. In fact, this structure is well known in extremal graph theory and is commonly referred to as the {\em reduced graph}.
Some important auxiliary results, such as the so-called Key Lemma \cite{KomSim96,Kom+02}
or the Blow Up lemma \cite{KomSarSze97}, reveal that this graph does inherit many of the essential structural properties of the original graph. 

%A regular partition reveals the existence of a hidden structure in a graph.
%Hence, despite their dissimilarities, it could be interesting to try to combine the two partitioning approaches in order to obtain a novel and efficient clustering strategy. In fact, extremal graph theory provides us with some interesting results and abstract structures that can be conveniently employed for our purpose: these are the notion of the {\em reduced graph} and the so-called Key Lemma \cite{????}. 

\begin{figure}[!t]
\centering
\includegraphics[scale=0.2]{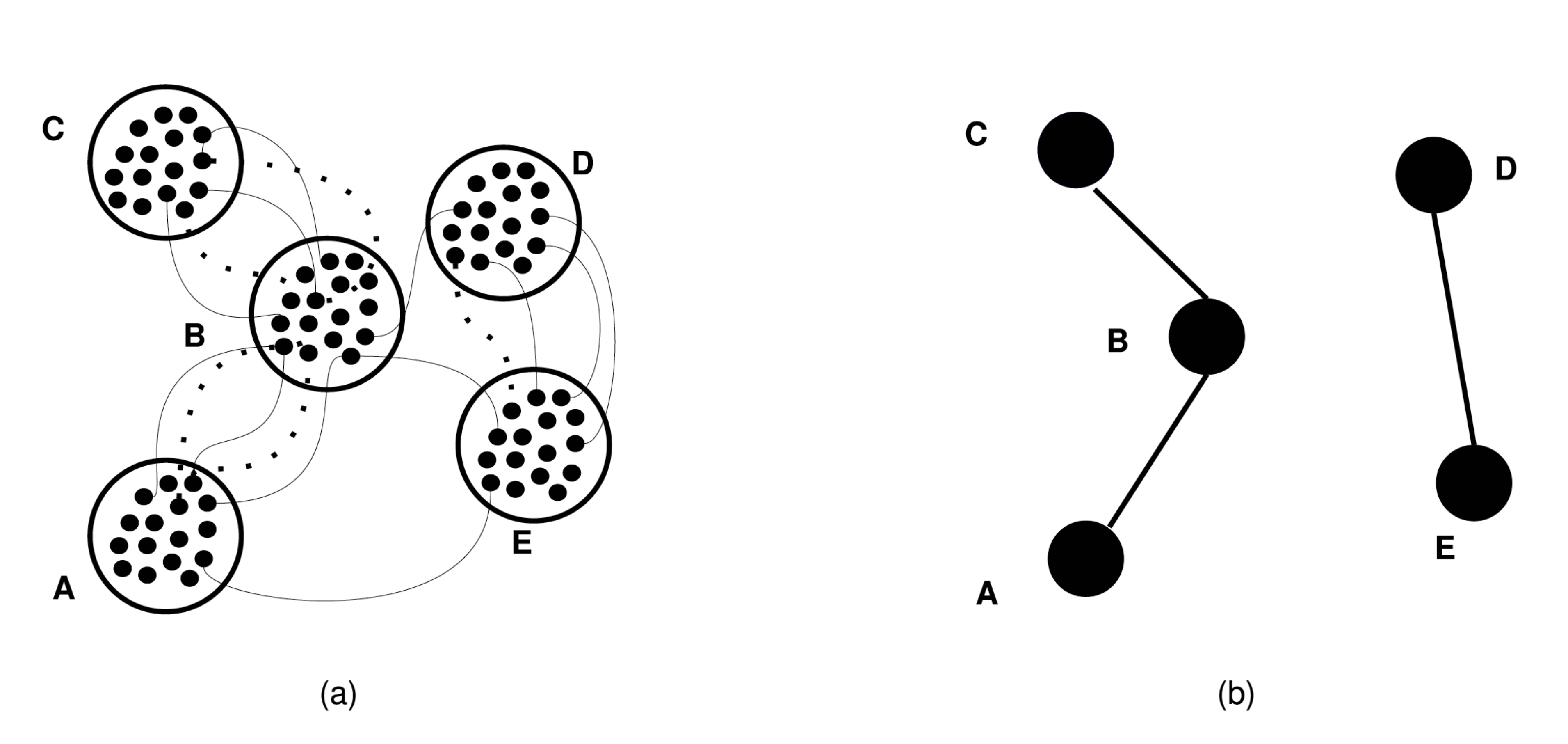}
\caption{Example of reduced graph creation. Left: original regular partition. Here, only the pairs $(A,B)$, $(B,C)$ and $(D,E)$ are regular. Right: the graph obtained after the reduction process.
From \cite{SpePel07}.
}
\label{redgraph}
\end{figure}

Given a graph $G = (V,E)$, a partition $P$ of the vertex-set $V$ into the sets $C_1, C_2, . . . C_k$ and two parameters $\varepsilon$ and $d$, the {\em reduced graph} $R$ is defined as follows. The vertices of $R$ are the clusters $C_1, C_2, . . .C_k$, and $C_i$ is adjacent to $C_j$ if
$(C_i, Cj)$ is $\varepsilon$-regular with density more than $d$. 
Figure \ref{redgraph} shows an example of transformation from a partitioned graph to its reduced graph. 
Consider now a graph $R$ and an integer $t$. 
Let $R(t)$ be the graph obtained from $R$ by replacing each vertex $x \in V(R)$ by a set $V_x$ of $t$ independent vertices, and joining $u \in V_x$ to $v \in V_y$ if and only if $(x, y)$ is an edge in $R$.
$R(t)$ is a graph in which every edge of $R$ is replaced by a copy of the complete bipartite graph $K_{tt}$.
The following Lemma shows how to use the reduced graph $R$ and its modification $R(t)$ to infer properties of a more complex graph.

\begin{theorem}[Key Lemma]
\label{KeyLemma}
Given the reduced graph $R$, $d > \varepsilon > 0$, a positive integer $m$, construct graph $G^*$ which is an approximation of graph G by performing these two steps:
\begin{enumerate}
\item replace every vertex of $R$ by $m$ vertices.
\item replace the edges of $R$ with regular pairs of density at least $d$.
\end{enumerate}
Let $H$ be a subgraph of $R(t)$ with $h$ vertices and maximum degree $\Delta >0$, and let $\delta = d-\varepsilon$ and $\varepsilon_0 = \delta^\Delta/(2+\Delta)$. If $\varepsilon \leq \varepsilon_0$ and $t-1 \leq \varepsilon_0m$, then $H$ is embeddable into $G$ (i.e., $G$ contains a subgraph isomorphic to $H$).
In fact, we have:
\begin{equation}
\left\vert\left\vert{H\,\to\,G}\right\vert\right\vert > (\varepsilon_0m)^h
\end{equation}
where $\left\vert\left\vert{H\,\to\,G}\right\vert\right\vert$ denotes the number of labeled copies of $H$ in $G$.
\end{theorem}

Given a graph $R$, the Key Lemma furnishes rules to expand $R$ to a more complex partitioned graph $G$ which respects edge-density bounds. On the other hand, we have another expanded graph, $R(t)$. Because of their construction, $R(t)$ and $G$ are very similar, but they can have a different vertex cardinality. In addition, note that the only densities allowed between vertex subsets in $R(t)$ are 0 and 1. The Key Lemma establishes constraints to the edge density d and the subset size $t$ in order to assure the existence of a fixed graph $H$ embeddable into $R(t)$, which is also a subgraph of G. Let $H$ be a subgraph of $R(t)$. 
If  \textit{t} is sufficiently small with respect to \textit{m}, and \textit{d} is sufficiently high with respect to $\varepsilon$ and $\Delta$, it is possible to find small subsets of vertices such that they are connected with a sufficiently high number of edges. The copies of $H$ are constructed vertex by vertex, by picking up elements from the previous identified subsets. 

\begin{figure}[!t]
\centering
\includegraphics[scale=0.22]{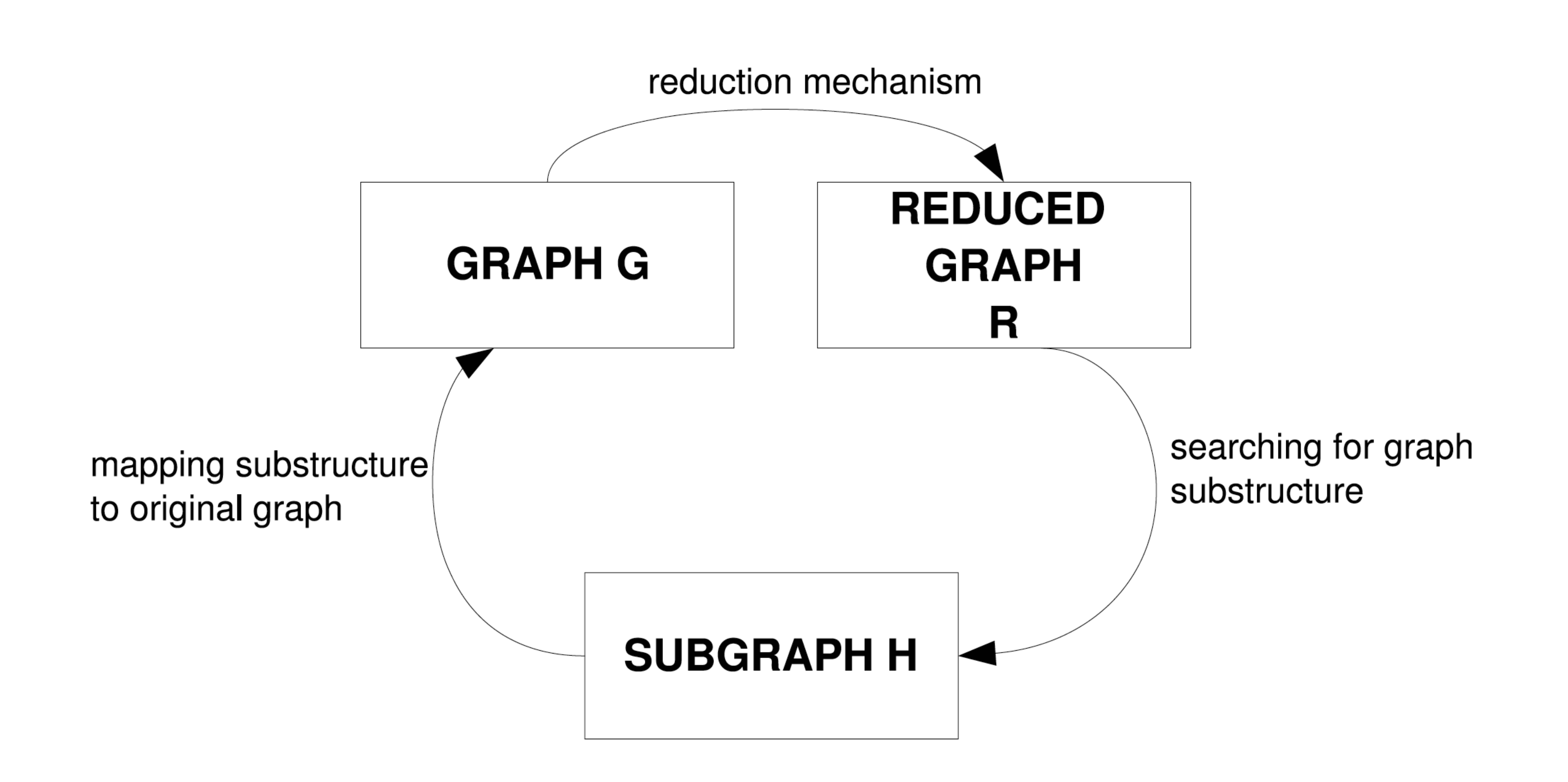}
\caption{Reduction strategy to find significant substructures in a graph.
From \cite{SpePel07}.
}
\label{2stage}
\end{figure}

As described in \cite{KomSim96}, a common and helpful combined use of the reduced graph
and the Key Lemma is as follows (see Figure \ref{2stage}):
\begin{itemize}

\item
Start with a graph $G = (V,E)$ and apply the regularity lemma, finding a regular partition $P$;

\item
Construct the reduced graph $R$ of $G$, w.r.t. the partition $P$;

\item
Analyze the properties of $R$, in particular its subgraphs;

\item
As it is assured by Theorem \ref{KeyLemma}, every small subgraph of $R$ is also a subgraph of $G$.

\end{itemize}

In summary, a direct consequence of the Key Lemma is that it is possible to search for significant substructures in a reduced graph $R$ in order to find common subgraphs of $R$ and the original graph.

Now, returning to the clustering problem, the approach developed in \cite{SpePel07} consists in a two-phase procedure. In the first phase, the input graph is decomposed into small pieces using Szemer\'edi's partitioning process and the corresponding (weighted) reduced graph is constructed, the weights of which reflect edge-densities between class pairs of the original partition. Next, a standard graph-based clustering procedure is run on the reduced graph and the solution found is mapped back into original graph to obtain the final groups.
Experiments conducted on standard benchmark datasets confirmed the effectiveness of the proposed approach both in terms of quality and speed.

Note that this approach differs from other attempts aimed at reducing the complexity of pairwise grouping processes, such as \cite{Ben+03,Fow+04,PavPel04},
as the algorithm performs no sampling of the original data but works instead on a derived structure which does retain the important features of the original one. 

The ideas put forward in \cite{SpePel07} were recently developed and expanded by \cite{Sar+12} and \cite{Nou15}, who confirmed the results obtained in the original paper.
\cite{Curado2015} have recently applied these ideas to improve the efficiency of edge detection algorithms. They compared the accuracy and the efficiency obtained using the regularity-based approach with that obtained with a combination of a factorization-based compression algorithm and quantum walks. They achieved a huge speed up, from an average of 2 hours for an image of $125 \times 83$ pixels ($10375$ vertices) to $2$ minutes with factorization-based compression and of $38$ seconds with regularity compression.

\section{An example application: Image segmentation}

To give a taste of how the two-phase strategy outlined in the previous section works, here we present some fresh experimental results on the problem of segmenting gray-level images.
Each image is abstracted in terms of an edge-weighted graph where vertices represent pixels and edge-weights reflect the similarity between pixels. As customary, the similarity between pixels, say $i$ and $j$, is measured as a function of the distance between the corresponding brightness values, namely, $w(i, j) = exp(-((I(i)-I(j))^2 /\sigma^2)$, where $I(i)$ is the normalized intensity value of pixel $i$ and $\sigma$ is a scale parameter.

We took twelve images from Berkeley's BSDS500 dataset \cite{Arbelaez2011} and, after resizing them 
to 81 $\times$ 121 pixels, we segmented them using two well-known clustering algorithms, namely 
Dominant Sets (DS) \cite{PavPel07} and Spectral Clustering (SC) \cite{Ng01onspectral}. 
The results obtained were then compared with those produced by the two-phase strategy,
which consists of first compressing the original graph using regular partitions and then using the
clustering algorithm (either DS of CS) on the reduced graph \cite{SpePel07}.

Two well-known measures were used to assess the quality of the corresponding segmentations, namely
the Probabilistic Rand Index (PRI) \cite{Unnikrishnan} and the Variance of Information (VI) \cite{Meila2005}. 
The PRI is defined as:
\begin{equation}
 PRI(S,\{G_k\}) = \frac{1}{T} \sum_{i<j} [c_{ij}p_{ij} + (1 - c_{ij})(1-p_{ij})]
\end{equation}
where $S$ is the segmentation of the test image, $\{G_k\}$ is a set of ground-truth segmentations, $c_{ij}$ is the event that pixels $i$ and $j$ have the same label, $p_{ij}$ its probability, and $T$ is the total number of pixel pairs. The PRI takes values in $[0,1]$, where $PRI = 1$ means that the test image segmentation matches the ground truths perfectly.

The VI measure is defined as:
\begin{equation}
 VI(S,S') = H(S) + H(S') - 2I(S,S')
\end{equation}
where $H$ and $I$ represent the entropy and mutual information between the test image segmentation $S$ and a ground truth segmentation $S'$, respectively. The VI is a metric which measures the distance between two segmentations. It takes values in $[0,log_2 n]$, where $n$ is the total number of pixels, and $VI=0$ means a perfect match.

\addtolength{\tabcolsep}{-0.5ex}  
\begin{figure*}[!ht]
\begin{center}
\begin{tabular}{cccc}
\includegraphics[scale= 0.7]{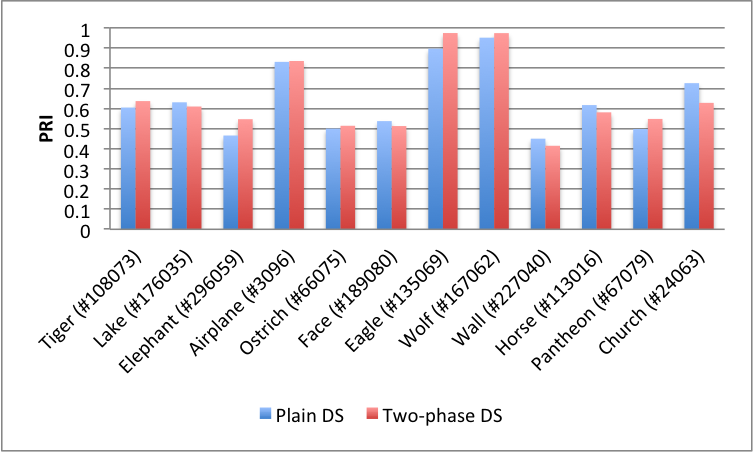} &
\includegraphics[scale= 0.7]{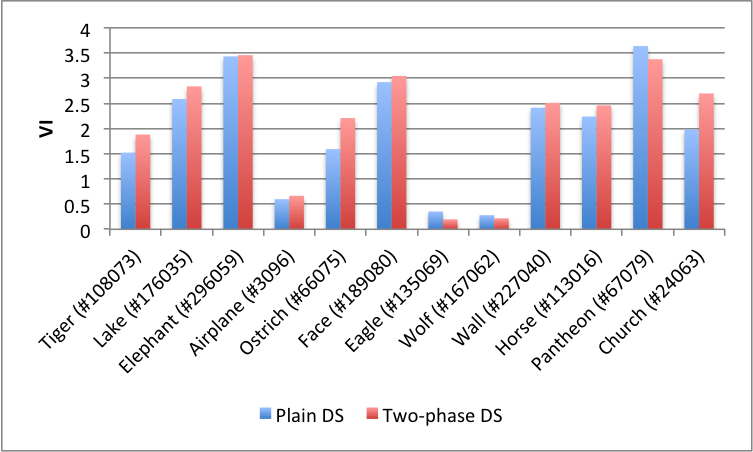} &\\		
\end{tabular}
\caption{Image segmentation results on twelve images taken from the Berkeley dataset using plain Dominant Sets (Plain DS) and the two-phase strategy described in Section \ref{sec:clustering} (Two-phase DS).
Left: Probabilistic Rand index (PRI). Right: Variance of Information (VI). (See text for explanation).
The numbers in parentheses on the x-axes represent the image identifiers within the dataset.}
\label{figure:DS-results}
\end{center}
\end{figure*}

\addtolength{\tabcolsep}{-0.5ex}  
\begin{figure*}[!ht]
\begin{center}
\begin{tabular}{cccc}
\includegraphics[scale= 0.7]{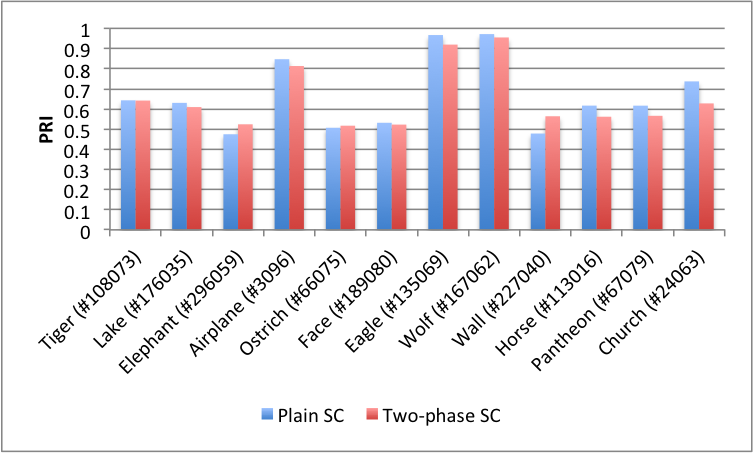} &
\includegraphics[scale= 0.7]{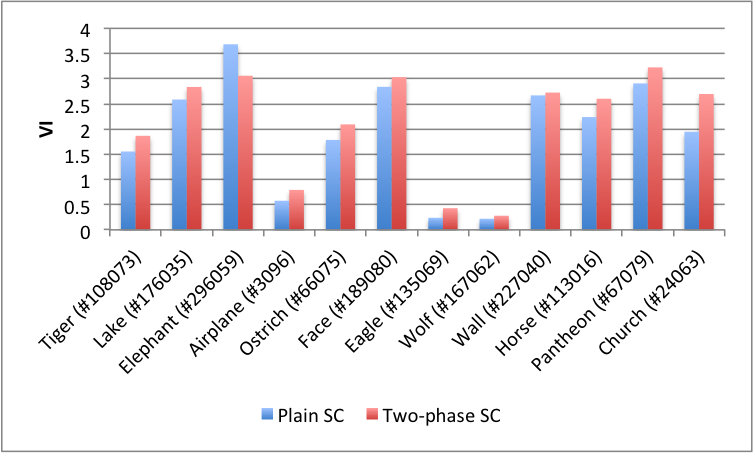} &\\		
\end{tabular}
\caption{Image segmentation results on twelve images taken from the Berkeley dataset using plain Spectral Clustering (Plain SC) and the two-phase strategy described in Section \ref{sec:clustering} (Two-phase SC).
Left: Probabilistic Rand index (PRI). Right: Variance of Information (VI). (See text for explanation).
The numbers in parentheses on the x-axes represent the image identifiers within the dataset.}
\label{figure:SC-results}
\end{center}
\end{figure*}

\begin{figure*}[!ht]
\begin{center}
\begin{tabular}{cccc}
\fbox{\includegraphics[scale= 0.8]{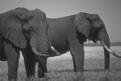}} &
\fbox{\includegraphics[scale= 0.8]{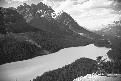}} &		
\fbox{\includegraphics[scale= 0.8]{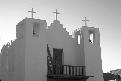}} & 
\fbox{\includegraphics[scale= 0.8]{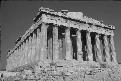}} \\[0.5ex]
\fbox{\includegraphics[scale= 0.8]{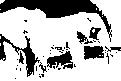}} &
\fbox{\includegraphics[scale= 0.8]{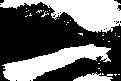}} &
\fbox{\includegraphics[scale= 0.8]{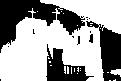}} &
\fbox{\includegraphics[scale= 0.8]{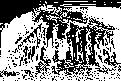}} \\[0.5ex]
\fbox{\includegraphics[scale= 0.8]{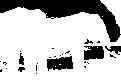}} &
\fbox{\includegraphics[scale= 0.8]{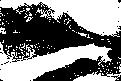}} &		
\fbox{\includegraphics[scale= 0.8]{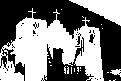}} & 
\fbox{\includegraphics[scale= 0.8]{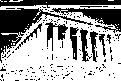}} \\ [0.5ex]
\fbox{\includegraphics[scale= 0.8]{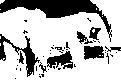}} &
\fbox{\includegraphics[scale= 0.8]{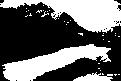}} &
\fbox{\includegraphics[scale= 0.8]{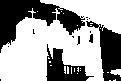}} &
\fbox{\includegraphics[scale= 0.8]{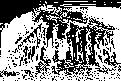}} \\ [0.5ex]
\fbox{\includegraphics[scale= 0.8]{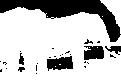}} &
\fbox{\includegraphics[scale= 0.8]{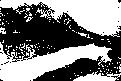}} &		
\fbox{\includegraphics[scale= 0.8]{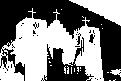}} & 
\fbox{\includegraphics[scale= 0.8]{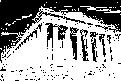}} \\ 
\end{tabular}
\caption{Comparing the segmentation results of plain clustering (DS/CS) and the 
two-phase approach.
First row: original images.
Second row: results of plain Dominant Set (DS) clustering. 
Third row: results of the two-phase Szemer\'edi+DS strategy.
Fourth row: results of plain Spectral Clustering (SC). 
Fifth row: results of the two-phase Szemer\'edi+SC strategy.
%Note that the size of the reduced graphs for the first three images is 4, while for the last image is 8.
}
\label{figure:visual}
\end{center}
\end{figure*}

The results are shown in Figures \ref{figure:DS-results} and \ref{figure:SC-results}, while
Figure \ref{figure:visual} shows the actual segmentations obtained for a few representative images.
Note that the results of the two-stage procedure
are comparable with those obtained by applying the corresponding clustering algorithms
directly on the original images, and sometimes are even better.
Considering that in all cases, the Szemer\'edi algorithm obtained a compression rate
well above 99\% (see Table \ref{table:CR} for details), this is in our opinion impressive.
Note that these results are consistent with those reported in 
\cite{SpePel07,Sar+12,Nou15,Curado2015}.

\addtolength{\tabcolsep}{-0.5ex}  

\begin{table}[t]
    \center
        \begin{tabular}{ ||c|c|c|c|| }
        \hline
        Image & ~~Original~~  & Size of       & ~~Compression~~  \\
              &  size     & ~~reduced graph~~ &  rate  \\
        \hline
        \hline
        Tiger  & 9801 & 16 & 99.84\% \\
        \hline
        Lake  & 9801 & 4 & 99.96\%\\
        \hline
        ~~Elephant  ~~  & 9801 & 64 & 99.35\% \\
        \hline
        Airplane  & 9801 & 16 & 99.84\% \\
        \hline
        Ostrich  & 9801 & 16 & 99.84\% \\
        \hline
        Face  & 9801 & 32 & 99.67\% \\
        \hline
        Eagle   & 9801 & 64 & 99.35\% \\
        \hline
        Wolf  & 9801 & 16 & 99.84\% \\
        \hline
        Wall  & 9801 & 8 & 99.92\% \\
        \hline
        Horse  & 9801 & 8 & 99.92\% \\
        \hline    
        Pantheon  & 9801 & 8 & 99.92\% \\
        \hline    
        Church  & 9801 & 4  & 99.96\% \\
        \hline    
        \end{tabular}
    \caption{Sizes of the reduced graphs after running the Szemer\'edi compression algorithm, and corresponding compression rates, for all images used.}
\label{table:CR}
\end{table}

\begin{table}[t]
    \center
    \begin{tabular}{||c | c | c | c | c||} 
    \hline
    Image & ~~$ind(P_1)$~~ & ~~$ind(P_2)$~~ & ~~i$nd(P_3)$~~ & ~~$ind(P_4)$~~ \\ 
    \hline\hline
    Tiger & 0.142  & 0.217 & 0.272 & 0.317\\ 
    \hline
    Lake & 0.004 & 0.085 & 0.129 & 0.173\\
    \hline    
    ~~Elephant~~ & 0.071 & 0.154 & 0.204 & 0.248\\ 
    \hline
    Airplane & 0.205  & 0.306 & 0.363 & 0.408\\ 
    \hline   
    Ostrich & 0.154   & 0.231 & 0.279 & 0.319\\ 
    \hline  
    Face & 0.006  & 0.061 & 0.102 & 0.135\\ 
    \hline   
    Eagle & 0.213 & 0.318  & 0.376 & 0.417\\ 
    \hline   
    Wolf & 0.014  & 0.103   & 0.181  & 0.215 \\ 
    \hline 
    Wall & 0.201   & 0.304  & 0.362   & 0.399  \\ 
    \hline   
    Horse & 0.078  & 0.169  & 0.214   & 0.252 \\ 
    \hline     
    Pantheon & 0.049 & 0.101 & 0.151 & 0.179\\
    \hline
    Church & 0.009 & 0.081 & 0.126 & 0.167\\
    \hline
    \end{tabular}
    \caption{Behavior of the index of partition (\ref{indexP}) in the first four steps of the Szemer\'edi compression stage, for all images used.}
    \label{table:index}
\end{table}

%We used four images for this experiment. On the first row, we show the original images. On the second row we show the results of the dominant set clustering on those images. On the third row we give the results of the two-phased strategy, when we first apply the regularity lemma of Szemeredi and then we perform dominant set clustering on the reduced graph. Finally on the next two rows, we do the same, but this time with the spectral clustering algorithm instead of the dominant set algorithm. The size of the reduced graph for the first three images is 4, while for the last image is 8.

%By visual inspection, we see that the results of the two-phased strategy are comparable with the plain clustering. In fact, for the first and last image, the two-phased algorithm seems to give visually nicer results than by just clustering the original image, while for the two other images, our results are almost as good as the results of plain clustering. This came to us as a surprise, considering that the reduced graph (of the images shown here) is extremely small, and so we have more than 99.9\% compression.

We also investigated the behavior of the index of partition $ind(P)$ defined in (\ref{indexP}),
during the evolution of the Szemer\'edi compression algorithm.
Remember that this measure is known to increase at each iteration of Alon et al.'s (exact) algorithm described in Section \ref{sec:algo}, and it is precisely this monotonic behavior which guarantees the correctness of the algorithm (and, in fact, of Szemer\'edi's lemma itself). With our heuristic modifications, however, there is no such guarantee, and hence it is reassuring to see that in all cases the index does in fact increase at each step, as shown in Table \ref{table:index}, thereby suggesting that the simple heuristics described in \cite{SpePel07,Sar+12} appear not to alter the essential features of the original algorithm. A similar behavior was also reported in \cite{Sar+12}.

\section{Related works}

Besides the use of the regularity lemma described above, in the past few years there have been other algorithms explicitly inspired by the notion of a regular partition which we briefly describe below.

\cite{Nepusz2008} introduced a method inspired by Szemer\'edi's regularity lemma to predict missing connections in cerebral cortex networks. To do so, they proposed a probabilistic approach where every vertex is assigned to one of $k$ groups based on its outgoing and incoming edges, and the probabilistic description of connections between and inside vertex groups are determined by the cluster affiliations of the vertices involved. In particular, they introduced a parametrized stochastic graph model, called the preference model, which is able to take into account the amount of uncertainty present in the data in terms of uncharted connections. Their method was tested on a network containing $45$ vertices and $463$ directed edges among them. The comparison of their experimental results with the original data showed that their algorithm is able to reconstruct the original visual cortical network with an higher accuracy compared to state-of-the-art methods.

These good results motivated \cite{Pehkonen2011} to develop a method inspired by the notion of regular partition to analyse an experimental peer-to-peer system. This network is modeled as directed weighted graph, where an edge direction indicates a client-server relation and a weight is the proportion of all chunks obtained from such link (edge) during the whole experiment. Their aim was to understand the peer's behavior. In particular, they want to group peers with similar behavior with respect to downloading and uploading characteristics in the same cluster. Their approach exploits max likelihood estimation to extract a partition where all cluster pairs are as much as possible random bipartite subgraphs. Their method was tested on a small network of $48$ vertices of a p2p experimental network. The results showed that their algorithm detected some hidden statistical properties of the network. They pointed out that for larger systems, sharper results could be expected, although for larger networks an algorithmic version of Szemer\'edi's Regularity Lemma could be more plausible solution.

More recently, Szemer\'edi's lemma inspired \cite{Reittu2014}, who developed a variant of stochastic block models \cite{Peixoto} for clustering multivariate discrete time series. To this end, they introduced a counterpart of Szemer\'edi regular partition, called regular decomposition, which is a partition of vertices into $k$ sets is such a way that structure between sets and inside sets are random-like.
In particular, the number of clusters $k$ increases at each iteration of their algorithm as long as large clusters are created. The stopping criterion is provided by means of Rissanen's minimum description length (MDL) principle. This choice is driven by the regularity lemma: the algorithm searches for large regular structure, corresponding to a local MDL optimum with the smallest value of $k$. The application of their method to real-life electric smart meter customer has given structures which are more informative than of the structures which are obtained by means of a traditional clustering method as $k$-means.

Finally, we mention the recent work of \cite{Bonchi2013} who have introduced a local algorithm for correlation clustering to deal with huge datasets. In particular, they took inspiration from the PTAS for dense MaxCut of \cite{Frieze2004} and used low-rank approximations to the adjacency matrix of the graph. The algorithm searches a weakly regular partition for the graph in sub-linear time to get a good approximate clustering. They pointed out that their algorithm could be naturally adapted to distributed and streaming systems to improve their latency or memory usage. Thus, it can be used to detect communities in large-scale evolving graphs.

\section{Summary and outlook}

Szemer\'edi's regularity lemma is a deep result from extremal graph theory which states that every graph can be well-approximated by the union of a constant number of random-like bipartite graphs, called regular pairs. 
This and similar results provide us with a principled way to obtain a good description of a large graph using a small amount of data, and can be regarded as a manifestation of the pervasive dichotomy between structure and randomness. Although the original proof of the lemma was non-constructive, polynomial-time algorithms and heuristics have been developed to determine regular (or approximately regular) partitions for arbitrary graphs. 
In this paper, after introducing Szemer\'edi's result and its algorithmic aspects,
we have presented a review of the emerging applications of these ideas to structural pattern recognition, in particular to the area of graph-based clustering and image segmentation.

We think that the notion of a regular partition and related structures deserve a special attention in the field of pattern recognition and machine learning. Indeed, we are confident that this will allow us to develop a sound computational framework to effectively deal with classical problems arising in the context of massive networked data. This is also motivated by the fact that despite the increasing popularity of graph-based methods over traditional (feature-based) approaches, a typical problem associated to these algorithms is their poor scaling behavior, which hinders their applicability to problems involving large and dynamic data sets. Our hypothesis is that the use of regular partitions will allow us to perform recognition and learning on a compactly represented (possibly evolving) data, by separating structure from randomness. This has the potential to open up an exciting era where modern graph theory and combinatorics are combined in a principled way with mainstream machine learning and data mining.

\section*{Acknowledgments}
We are grateful to M. Bolla, H, Reittu and F. Bazs\'o for reading a preliminary version of the paper, and to the anonymous reviewers
for their constructive feedback.

\bibliographystyle{IEEEtran}
\balance
\bibliography{paper}

\end{document}